\documentclass[letterpaper, 10 pt, conference]{ieeeconf}  % Comment this line out if you need a4paper

\usepackage{url,graphicx,tabularx,array,verbatim,multirow,epstopdf,amsmath,amssymb,mathrsfs,bm,hyperref}
\usepackage[hypcap,tableposition=top]{caption}
\usepackage{subcaption}
\usepackage{float}
\floatstyle{plaintop}
\restylefloat{table}
\usepackage[english]{babel}
\usepackage{tikz}
\usetikzlibrary{spy,calc,external}
\usepackage{pgfplots}
\pgfplotsset{compat=newest}
\usepackage{pgfplotstable}
\pgfplotsset{every axis/.append style={font=\footnotesize}}
\newcommand{\subparagraph}{}
\usepackage{titlesec}
\usepackage{filemod}

\restylefloat{table}

\IEEEoverridecommandlockouts                              % This command is only needed if 
\renewcommand{\baselinestretch}{0.99}
\titlespacing*{\section}{0pt}{0.3\baselineskip}{0.3\baselineskip}
\titlespacing*{\subsection}{0pt}{0.3\baselineskip}{0.3\baselineskip}
\title{\LARGE \bf
Unsupervised Contact Learning for Humanoid Estimation and Control}

\author{Nicholas Rotella$^{1}$, Stefan Schaal$^{1,2}$ and Ludovic Righetti$^{2,3}$% <-this % stops a space
\thanks{This research was supported in part by National Science Foundation grants IIS-1205249, IIS-1017134, EECS-0926052, the Office of Naval Research, the Okawa Foundation, the Max-Planck-Society and the European Research Council under the European Union's Horizon 2020 research and innovation program (grant agreement No 637935). Any opinions, findings, and conclusions or recommendations expressed in this material are those of the author(s) and do not necessarily reflect the views of the funding organizations.}% <-this % stops a space
\thanks{$^{1}$Computational Learning and Motor Control Lab, University of Southern California, Los Angeles, California.}% <-this % stops a space
\thanks{$^{2}$Autonomous Motion Department, Max Planck Institute for Intelligent Systems, Tuebingen, Germany.}% <-this % stops a space 
\thanks{$^{2}$New York University, New York, New York}% <-this % stops a space 
}% <-this % stops a space

\begin{document}

\maketitle
\thispagestyle{empty}
\pagestyle{empty}

%%%%%%%%%%%%%%%%%%%%%%%%%%%%%%%%%%%%%%%%%%%%%%%%%%%%%%%%%%%%%%%%%%%%%%%%%%%%%%%%
\begin{abstract}
This work presents a method for contact state estimation using fuzzy clustering to learn contact probability for full, six-dimensional humanoid contacts.  The data required for training is solely from proprioceptive sensors - endeffector contact wrench sensors and inertial measurement units (IMUs) - and the method is completely unsupervised.  The resulting cluster means are used to efficiently compute the probability of contact in each of the six endeffector degrees of freedom (DoFs) independently.  This clustering-based contact probability estimator is validated in a kinematics-based base state estimator in a simulation environment with realistic added sensor noise for locomotion over rough, low-friction terrain on which the robot is subject to foot slip and rotation.  The proposed base state estimator which utilizes these six DoF contact probability estimates is shown to perform considerably better than that which determines kinematic contact constraints purely based on measured normal force.
\end{abstract}

%%%%%%%%%%%%%%%%%%%%%%%%%%%%%%%%%%%%%%%%%%%%%%%%%%%%%%%%%%%%%%%%%%%%%%%%%%%%%%%%
\section{INTRODUCTION}

Control and estimation approaches for legged robots rely on assumptions about the contact state of the feet.  Floating-base inverse dynamics resolves underactuation by projecting the dynamics into the contact constraints, forcing the endeffector acceleration to be zero \cite{mistry_inverse_2010}.  Locomotion on rough terrain focuses on stabilization through footstep adaptation but ignores the difficulties presented by contact constraint violations \cite{barasuol_reactive_2013}.  Similarly, legged odometry for base state estimation assumes the pose of an endeffector in contact is constant \cite{bloesch_state_2012}. Methods have been introduced to robustify kinematics-based base state estimation, including computing a contact point with minimal instantaneous velocity \cite{masuya_dead_2013} and outlier detection to discard measurements during slip \cite{bloesch_state_2013}, however few consider relaxing the contact assumptions by estimating contact quality in parallel.

Contact estimation is a broad topic which has been investigated in various contexts.  Petrovskaya et al.\cite{petrovskaya_probabilistic_2007} were among the first to consider multi-contact force control scenarios in which a manipulator interacts with the environment at points other than its endeffector.  Del Prete et al. investigated the effect of contact point estimation error on force control for a humanoid with a capacitive skin \cite{delprete_control_2012}.  Similar work was done by Manuelli et al. \cite{manuelli_localizing_2016} to estimate contact points without a tactile skin by fusing proprioceptive sensing with the dynamic model.

While estimation of contact points has been thoroughly studied, the problem of determining the \emph{quality} of contacts is less well-defined. One aspect of contact quality is the determination of contact constraint directions for a given task.  Ortenzi et al. \cite{ortenzi_kinematics_2016} computed endeffector constraints for a manipulator in contact with a surface using only kinematics, and Nozawa et al. \cite{nozawa_online_2017} recently presented a similar method for estimating environment constraints in humanoid control tasks.

For humanoids, the quality of a contact is largely determined by friction.  Hoepflinger et al. \cite{hoepflinger_unsupervised_2013} investigated foothold quality using unsupervised learning.  Terrain elevation map samples were clustered to find a set of primitives which were evaluated for foothold robustness by computing the friction coefficient through exploratory force control.  This allowed for prediction of contact quality from visual features for planning.  While Focchi et al. \cite{focchi_high_2017} employed offline friction estimation through for a quadruped walking on steep slopes, Ridgewell et al. \cite{ridgewell_humanoid_2017} introduced methods for online friction estimation and control adaptation.  While the friction coefficient determines linear slip, the center of pressure (CoP) boundaries determine rotational slip/roll.  Most controllers assume that the support polygon is the same shape as the foot, however this is invalid on rough terrain where line and even point contacts are encountered.  Wiedebach et al. \cite{wiedebach_walking_2016} presented one of the only approaches for online CoP boundary estimation during terrain exploration.

In contrast to approaches which indirectly compute contact quality by computing friction and CoP bounds, we wish to avoid contact models by directly estimating the probability of an endeffector being constrained in each of its six DoFs independently.  In this direction, Hwangbo et al. \cite{hwangbo_probabilistic_2016} developed a Hidden Markov Model which uses kinematic and dynamic models to predict contact transitions without force sensing.  This one-dimensional approach requires little sensing, however it does not estimate contact quality of the contact nor does it evaluate the classifier in an estimator.  

Camurri et al. \cite{camurri_probabilistic_2017} recently developed a method for contact probability estimation using logistic regression.  This one-dimensional classifier learns the normal force threshold at which the contact state transitions, ignoring lateral forces under the assumption that sufficient friction exists to prevent slip.  The resulting probability per endeffector is used to weight the corresponding measurement in a base state estimator, and a heuristic for modulating the measurement variance to filter out the effect of impacts is introduced. Although this estimator performs better than one using a fixed threshold, the classifier requires significant effort to train - ground truth is obtained manually as the contact sequence which minimizes estimation error.  Further, all results shown are for walking on flat ground where slipping does not occur.  Finally, only one dimension (normal force) is considered.

In contrast, we develop a contact estimator which:
\begin{itemize}
\item is completely unsupervised and model-free
\item uses only common, proprioceptive sensors (endeffector force/torque and IMU) 
\item estimates the probability of contact in all six endeffector DoFs independently
\end{itemize}

We test this contact estimator for use in base state estimation by modulating the measurement uncertainty associated with each endeffector DoF using the corresponding estimated probability of contact.  The following section details the motivation and setup for this approach.

\section{BACKGROUND}

\subsection{Motivation}

A difficult question arises when designing an estimator for contact state: what does it mean for an endeffector to be \emph{in contact}?  Most approaches treat the endeffector as fixed to contact surface if the normal force exceeds a chosen threshold, however contact truly occurs when the assumed endeffector constraints are satisfied - these statements are not always the same. The six DoF endeffector constraints are equivalent to enforcing that the feet cannot \emph{slip} or \emph{rotate}.  An endeffector will not slip if the static friction constraint
\begin{equation}
\label{eq:friction}
\sqrt{F_{x}^{2}+F_{y}^{2}} \leq \mu_{x,y}F_{z}
\end{equation}
is satisfied, where $F$ is the contact force and $\mu_{x,y}$ is the translational coefficient of friction.  Likewise, the endeffector will not rotate if the CoP and rotational friction constraints
\begin{align}
\begin{bmatrix}
-\tau_{y}/F_{z}\\
 \tau_{x}/F_{z}
\end{bmatrix} &\leq
\begin{bmatrix}
CoP_{x}\label{eq:cop}\\
CoP_{y}
\end{bmatrix}\\
\left|\tau_{z}\right| &\leq \mu_{z}F_{z}\label{eq:rot_friction}
\end{align}
are satisfied, where $\tau$ is the contact torque, $\mu_{z}$ is the rotational coefficient of friction and $CoP_{x}$, $CoP_{y}$ denote the contact support polygon bounds which are functions of contact surface geometry.

Since a sufficiently-high normal force $F_{z}$ would guarantee that inequalities (\ref{eq:friction}-\ref{eq:rot_friction}) are satisfied regardless of the other contact wrench dimensions, most estimation approaches simply threshold $F_{z}$ \cite{bloesch_state_2012}, \cite{fallon_lidar_2014}, \cite{faraji_practical_2015}.  However, this is restrictive especially on rough terrain where low friction and difficult surface geometry make slip and rotation likely even at high normal force values.  It also results in a one-dimensional contact state estimate as in \cite{hwangbo_probabilistic_2016}, \cite{camurri_probabilistic_2017}, whereas the contact constraint is truly six-dimensional.

\subsection{Sensing for Clustering}
\label{sec:sensing_for_clustering}
As discussed in the previous section, contact constraints are invalid when an endeffector slips and/or rotates, which is caused by a violation of friction and/or CoP constraints; these constraints depend on the contact wrench and surface properties (friction coefficients and geometry).  Rather than estimate these properties, we seek to cluster measured contact wrench data to directly learn constraint probabilities.   

All experiments in this work are performed in the SL simulation environment \cite{schaal_sl_2007}; we add simulated random-walk biases $b_{F}$ and $b_{\tau}$, along with simulated Gaussian noise processes $w_{F}$ and $w_{\tau}$, to the true force $F$ and torque $\tau$ measurements:
\begin{align}
\label{eq:ft}
F &= \bar{F} + b_{F} + w_{F}\\
\tau &= \bar{\tau} + b_{\tau} + w_{\tau}
\end{align}
As low-cost IMUs become available, humanoids are being augmented with additional sensing to improve estimation \cite{rotella_inertial_2016}, \cite{xinjilefu_imus_2016}; in order to give structure to the clustering problem, we add a simulated IMU to each endeffector.  We model the sensor outputs subject to simulated random-walk biases and thermal noise processes \cite{woodman_inertial_2007} as
\begin{align}
\label{eq:imu}
a^{IMU} &= R_{W}^{IMU}(a^{W}+g) + b_{a} + w_{a}\\
\omega^{IMU} &= R_{W}^{IMU}\omega^{W} + b_{\omega} + w_{\omega}
\end{align}
where $a\in R^{3}$ and $\omega\in R^{3}$ are the linear acceleration and angular velocity, respectively.  $R_{W}^{IMU}\in SO(3)$ is the rotation from world to IMU frame and $g$ is gravity. Sensors are assumed to be aligned with the endeffector frame, however their positions relative to this frame origin are not required.

\begin{table}[h!]
\centering
\begin{tabular}{|c|c|c|}
\hline
 & Continuous & Discrete ($1kHz$)\\ \hline
$\sigma_{\theta}$ & $0.00000316 rad / \sqrt{Hz}$ & $0.0001 rad$\\ \hline
$\sigma_{F}$ & $0.06325 N / \sqrt{Hz}$ & $2 N$\\ \hline
$\sigma_{b_{F}}$ & $0.0001 N / s / \sqrt{Hz}$ & $0.00316 N / s$\\ \hline
$\sigma_{\tau}$ & $0.00316 Nm / \sqrt{Hz}$ & $0.1 Nm$\\ \hline
$\sigma_{b_{\tau}}$ & $0.0001 Nm / s / \sqrt{Hz}$ & $0.00316 Nm / s$\\ \hline
$\sigma_{a}$ & $0.00078 m / s^{2} / \sqrt{Hz}$ & $0.02467 m / s^{2}$\\ \hline
$\sigma_{b_{a}}$ & $0.0001 m / s^{3} / \sqrt{Hz}$ & $0.00316 m / s^{3}$\\ \hline
$\sigma_{\omega}$ & $0.000523 rad / s / \sqrt{Hz}$ & $0.01653 rad / s$\\ \hline
$\sigma_{b_{\omega}}$ & $0.000618 rad / s^{2} / \sqrt{Hz}$ & $0.01954 rad / s^{2}$\\
\hline
\end{tabular}
\caption{Simulated sensor noise standard deviations. Corresponding values for 1kHz sampling rate are shown.}
\label{table:noise}
\end{table}

\section{CLUSTERING SETUP}
\label{sec:clustering}

Because we seek a continuous measure of contact quality rather than a classifier, we employ Fuzzy C-means (FCM) clustering which results in the soft partitioning of a dataset by allowing each data point to belong to more than one cluster \cite{dunn_fuzzy_1974}.  This is accomplished by minimizing the cost

\begin{equation}
\label{eq:cmeans_cost}
\sum_{i=1}^{N_{p}}\sum_{j=1}^{N_{c}}w_{i,j}^{m}||x_{i}-c_{j}||^{2},\quad m > 1
\end{equation}

where $N_{p}$ is the number of data points $x_{i}$, $N_{c}$ is the chosen number of clusters, $w_{i,j}$ is the membership weight of point $i$ belonging to cluster $j$ and $m$ is a constant which can be used to tune the amount of cluster overlap.  

This cost is minimized in a manner similar to k-means clustering; first, initial membership weights are randomly assigned.  Then, cluster means are computed as

\begin{equation}
\label{eq:cmeans_mean}
c_{j} = \frac{\sum_{i=1}^{N_{p}}w_{i,j}^{m}x_{i}}{\sum_{i=1}^{N_{p}}w_{i,j}^{m}}
\end{equation}

after which new membership weights are computed with

\begin{equation}
\label{eq:cmeans_weight}
w_{i,j} = \frac{1}{\sum_{k=1}^{N_{c}}\left(\frac{||x_{i}-c_{j}||}{||x_{i}-c_{k}||}\right)^{\frac{2}{m-1}}}
\end{equation}

Eq. (\ref{eq:cmeans_mean}-\ref{eq:cmeans_weight}) are iterated until the membership weights converge.  Since $\sum_{j=1}^{N_{c}}w_{i,j}=1$, we treat $w_{i,j}$ as the probability of point $i$ belonging to cluster $j$.

We use an FCM implementation from the Python library Scikit-learn \cite{scikit-learn} with $N_{C}=2$ clusters (corresponding to \emph{contact} and \emph{no contact} states) and default stopping parameters.  The ``fuzziness'' constant is set to $m=1.2$ which is the default value in most libraries.  Increasing this factor can amplify the effect of slip on contact probability, however it also reduces the probability of contact when no slip occurs.

Each data point $x_{k}\in R^{7T}$ is a time series of the past $T=20$ samples (at our control rate, $0.020s$). We include a short time-history to improve estimation response time; optimization of this time window is left to future work.  

Clustering is performed independently for the six DoF $\{x,y,z,\alpha,\beta,\gamma\}$ of each endeffector, using the full contact wrench and the corresponding IMU dimension from $\{a^{IMU}_{x},a^{IMU}_{y},a^{IMU}_{z},\omega^{IMU}_{x},\omega^{IMU}_{y},\omega^{IMU}_{z}\}$.  The constraint in the local endeffector frame $y$ direction uses, for example, data points of the form
\begin{align}
\label{eq:cluster_data}
x_{k} &=\\
 \{&\{F_{x_{k-T}},\cdots,F_{x_{k}}\}, \{F_{y_{k-T}},\cdots,F_{y_{k}}\}, \{F_{z_{k-T}},\cdots,F_{z_{k}}\},\nonumber\\
&\{\tau_{x_{k-T}},\cdots,\tau_{x_{k}}\}, \{\tau_{y_{k-T}},\cdots,\tau_{y_{k}}\}, \{\tau_{z_{k-T}},\cdots,\tau_{z_{k}}\},\nonumber\\
&\{a^{IMU}_{y_{k-T}},\cdots,a^{IMU}_{y_{k}}\}\nonumber
\}
\end{align}
Data from sensors with noise added as in Sec. \ref{sec:sensing_for_clustering} is collected and used unfiltered for clustering.  Preprocessing entails dimension-wise normalization of all $x_{k}$ (to ensure that the scale of dimensions such as $F_{z}$ do not dominate) followed by taking the absolute value (since slip is bi-directional). 
\begin{comment}
In theory, it should not be necessary to use all six contact wrench DoFs for each clustering problem - for example, since foot rotation doesn't depend on the lateral forces $F_{x}$ and $F_{y}$ - however we use them all for simplicity.  If computation time is a concern then it should be possible to reduce the dimensionality of the clustering problem; however, this would also assume that the IMU is aligned with the force/torque sensor frame.  We leave this discussion to future work.
\end{comment}

\section{BASE STATE ESTIMATION}

In order to evaluate the utility of the proposed contact estimator, we incorporate it into a base state estimation framework which relies on stationary contact assumptions. In previous work \cite{rotella_state_2014} we have implemented a kinematics-based estimator which fuses IMU data and relative base pose measurements to estimate the floating base state of a humanoid.  The estimator measurements take the form
\begin{align}
\label{eq:bse_meas}
s_{p,i} &= R(q)(p_{i}-r) + n_{p}\\
s_{z,i} &= \exp(n_{z})\otimes q\otimes z_{i}^{-1}
\end{align}
where $R(q)$ denotes the rotation matrix corresponding to the estimated base quaternion $q$, $p_{i}$ and $z_{i}$ are the estimated foot $i$ position and quaternion respectively, and $n_{p}$ and $n_{z}$ are position and orientation measurement noise vectors (see \cite{rotella_state_2014} for more details).  In most approaches for legged robots, the variances of $n_{p}$ and $n_{z}$ are set to constant, tuned values and the measurements are dropped from the filter when the endeffector loses contact, determined based on a fixed normal force threshold.

In contrast, in this work we set the contact state (which determines active measurements) and measurement noise variance using the output of the probability estimator.  When the probability of contact vector $P_{contact}\in R^{6}$ exceeds $P_{i} = 0.5$ in every dimension $i$, we consider the endeffector \emph{in contact} and use the corresponding measurements.  Further, we set the measurement noise covariance matrix as
\begin{align} 
\label{eq:meas_cov}
\Sigma &= E[nn^{T}] = r^{2}I + \alpha(I - \mbox{diag}(P_{contact}))
\end{align}
where $r$ is the nominal measurement noise standard deviation (sometimes tuned separately for position and orientation) and $\alpha$ is a scaling factor for the probability-dependent term (we choose $\alpha=1$ for simplicity).  The covariance thus converges to its constant value as in \cite{rotella_state_2014} when $P_{contact}\rightarrow 1$. This is conceptually similar to the approach of \cite{camurri_probabilistic_2017} but requires less tuning and considers all six contact dimensions.

Since clustering is performed in the endeffector frame, the covariance must be transformed into the base frame where the base state estimator measurement is expressed. This is accomplished with
\begin{equation*}
\hat{\Sigma} = R\Sigma R^{T},\quad R = \mbox{blockdiag}(R_{Endeff}^{Base}, R_{Endeff}^{Base})\in R^{6\times 6}
\end{equation*}
where $R_{Endeff}^{Base}\in R^{3\times 3}$ is the rotation from endeffector to base frame (a function of kinematics and joint angles only).

\begin{figure*}[t]
\includegraphics[width=1.0\textwidth]{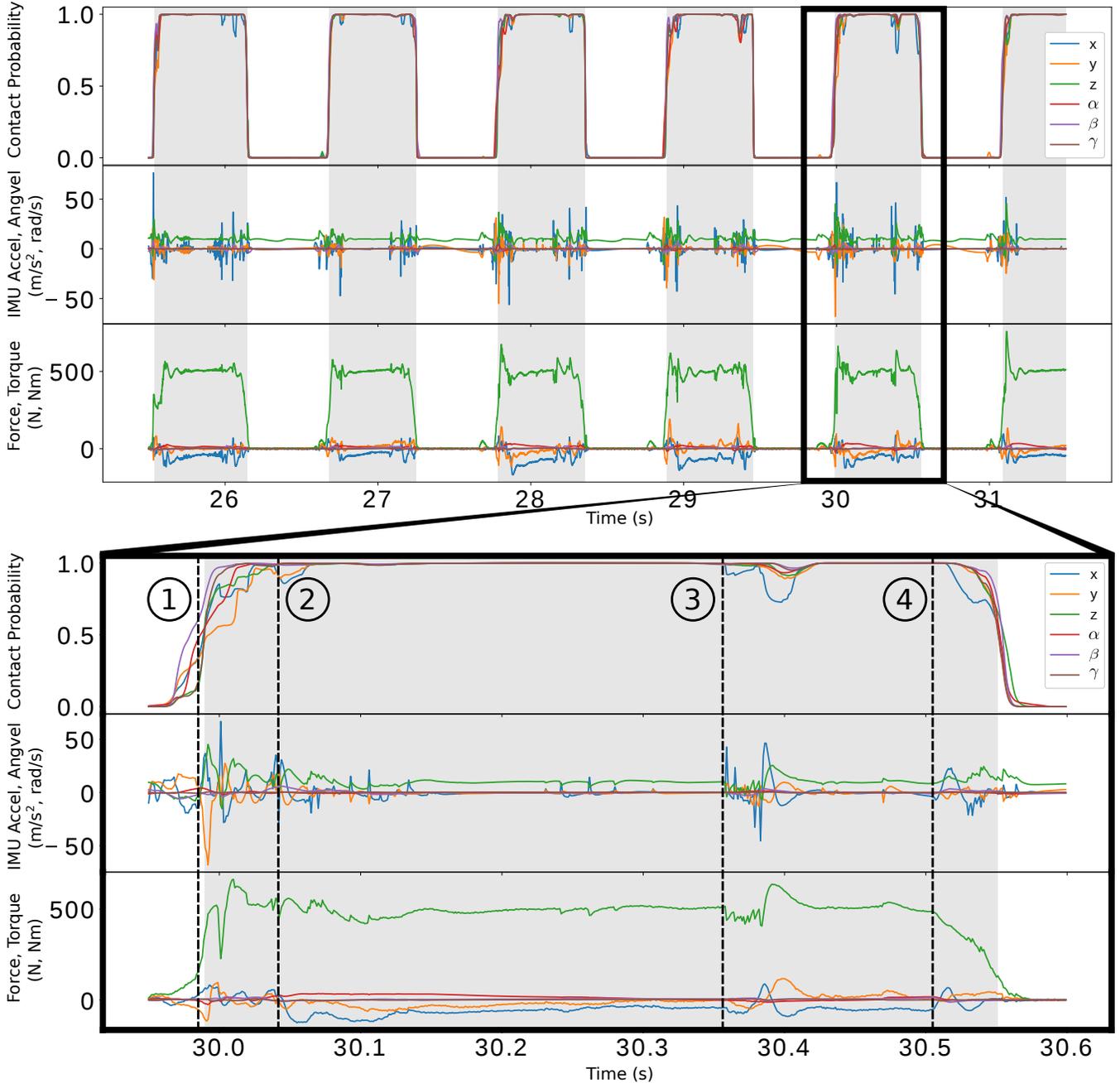}
\captionsetup{width=1.0\textwidth}
\caption{The top portion shows the six-dimensional contact probability resulting from a rough terrain walking task (top) along with the measured IMU linear acceleration and angular velocity (middle) and measured contact force and torque (bottom).  The portions in gray denote contact according to the probability estimator (all $P_{i}>0.5$).  The lower portion of the plot shows a zoomed view of one contact cycle with several distinctive contact events highlighted for discussion in Sec. \ref{sec:clustering_results}.}
\label{fig:contact_prob_fig}
\end{figure*}

\section{EXPERIMENTS AND RESULTS}

We perform a number of experiments to evaluate the performance of the proposed estimator and analyze its properties.  All experiments are performed in the SL simulator \cite{schaal_sl_2007} during a 60 second rough terrain walking task with simulated joint angle, IMU and contact wrench sensor noise as in Table \ref{table:noise}; noisy data is used for clustering, contact estimation and base state estimation.  Control is computed using non-noisy sensor data and ideal base state estimation, however we investigate using the proposed contact estimator for closed-loop control in Sec. \ref{sec:clustering_for_control}.  Walking velocity commands were recorded from user input and played back, producing repeatable trajectories across experiments.  The rough terrain consists of raised patches with a friction coefficient of $0.4$ (half the normal friction in SL).  To account for the effect of noise and slight contact differences, Root-Mean-Squared Error (RMSE) for experiments in this section was computed by averaging performance across ten trials.

\subsection{Contact Clustering Results}
\label{sec:clustering_results}

Sensor data was recorded from the rough terrain walking task and clustered offline as detailed in Sec. \ref{sec:clustering} (clustering takes on the order of a few seconds).  The cluster means were then used to compute contact probability during a similar walking task; the results are shown in Fig. \ref{fig:contact_prob_fig}.

We focus on one contact cycle in the lower portion of Fig. \ref{fig:contact_prob_fig} to investigate the clustering results more closely. Slip first occurs in the $y$ direction at $(1)$, causing the corresponding contact probability to lag behind the other dimensions.  Rotational slip in $\alpha$ is also present during loading, however on a smaller scale.  Slip then occurs in $x$ because the foot is not sufficiently loaded while the robot tries to create force in $-x$ to decelerate the center of mass; once $F_{z}$ increases, there is sufficient friction to stop slipping and a negative $F_{x}$ is sustained from (2) on.  A drop in $F_{z}$ during single support at $(3)$ again causes slip, leading to a decrease in contact probability in all dimensions.  Finally, slip in $x$ again occurs at $(4)$ as the foot is being unloaded.  These are only a few highlights of the complex contact interaction shown, however they aid in understanding where/why slip can occur.

\subsection{Base State Estimation Threshold}
\label{sec:bse_force_thresh}

\begin{figure}[h]
\includegraphics[width=0.49\textwidth]{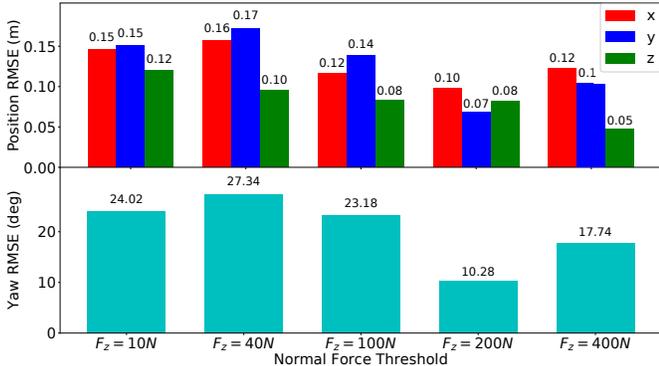}
\captionsetup{width=0.5\textwidth}
\caption{Root Mean Squared Error (RMSE) for estimation of the unobservable base position (top) and yaw (bottom) for different normal force thresholds.}
\label{fig:bse_force_thresh}
\end{figure}

Since we evaluate the proposed contact probability estimator against a typical humanoid base state estimator with a fixed normal force threshold for contact \cite{rotella_state_2014}, we first perform experiments to optimize the chosen threshold.  Performance is evaluated by computing the RMSE for the base position and yaw angle as these four states are always unobservable without adding exteroceptive sensing.  The normal force thresholds $\{10N, 40N, 100N, 200N, 400N\}$ were tested, with performance averaged across ten trials each; the results are shown in Fig. \ref{fig:bse_force_thresh}.  The RMSE mostly decreased for increasing thresholds, with $200N$ resulting in the best performance.  A threshold of $400N$ removes the double support period from estimation entirely, resulting in more error.  We use a fixed threshold of $200N$ for the baseline estimator in experiments in the remainder of this work. 

\subsection{Clustering-Based State Estimation}
\label{sec:cluster_vs_fixed}

\begin{figure}[h]
\includegraphics[width=0.49\textwidth]{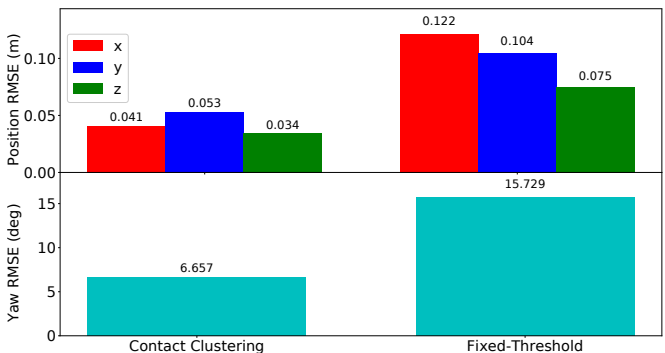}
\captionsetup{width=0.5\textwidth}
\caption{Root Mean Squared Error (RMSE) for estimation of the unobservable base position (top) and yaw (bottom) for the contact probability-based base state estimator and the fixed normal force threshold base state estimator.}
\label{fig:cluster_vs_fixed}
\end{figure}

We evaluate the base state estimator detailed in our previous work using both a fixed normal force threshold (as is commonly done) and using the proposed clustering-based contact probability estimator for the same rough terrain walking task.  The base state estimators are identical other than the measurement noise covariance matrix modulation of Eq. (\ref{eq:meas_cov}).  As shown in Fig. \ref{fig:cluster_vs_fixed}, using the contact clustering for base state estimation considerably reduces the RMSE.

\subsection{Clustering Training Data}
\label{sec:walk_vs_rough_rmse}

\begin{figure}[h!]
\includegraphics[width=0.49\textwidth]{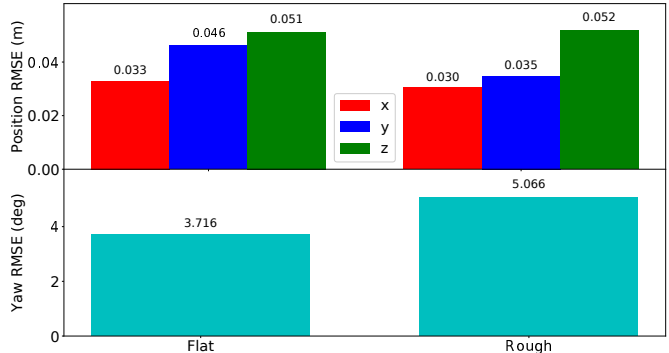}
\captionsetup{width=0.5\textwidth}
\caption{Root Mean Squared Error (RMSE) for estimation of the unobservable base position (top) and yaw (bottom) for different training datasets.}
\label{fig:walk_vs_rough_rmse}
\end{figure}

In order to test how well the clustering-based estimator generalizes to different types of terrain, we performed clustering using data from two different tasks: one which walks over \emph{rough} terrain (as in all other experiments) and one which walks in place on \emph{flat} ground.  We then tested both contact estimators with separate base state estimators on the same rough terrain walking task; the resulting estimation errors are shown in Fig. \ref{fig:walk_vs_rough_rmse}.

Surprisingly, the estimator trained on flat ground walking data performs roughly equally-well, despite having been trained on a much different dataset than was used for testing.  This is a desirable characteristic because obtaining data from rough terrain walking on a real robot is difficult, especially without accurate state estimation already in place.

\begin{figure}[h!]
\includegraphics[width=0.49\textwidth]{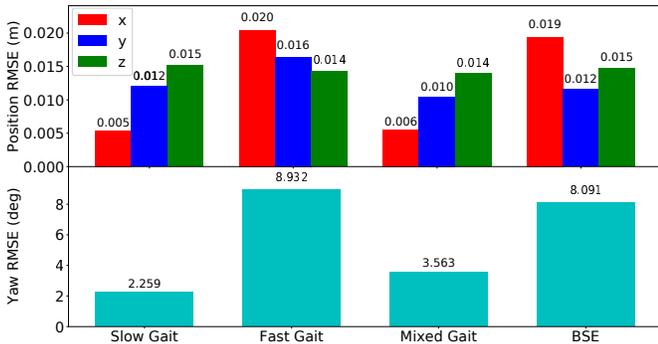}
\captionsetup{width=0.5\textwidth}
\caption{Root Mean Squared Error (RMSE) for estimation of the unobservable base position (top) and yaw (bottom) for walking in place on a patch of rough terrain with a varying gait using clustering trained on three different gait types as well as for the fixed-threshold base state estimator (BSE).}
\label{fig:gait_test_rough_rmse}
\end{figure}

We also wish to test how well the clustering-based estimator generalizes to different gaits.  We perform clustering using data from flat ground walking in varying directions using three different gaits.  The default gait used for walking in this work has a single support period of $0.5s$ and a double support period of $0.05s$; we denote this the \emph{fast} gait.  We also perform clustering on \emph{slow} gait data (single support period of $1.0s$, double support period of $0.5s$).  Finally, we cluster using data from a \emph{mixed} gait which varies throughout the task between fast and slow.  We then test the clustering-based estimators for these gaits for a mixed gait walk-in-place task on a pacth of rough terrain (varying the gait during a normal walking task over rough terrain is too unstable).  The results are shown in Fig. \ref{fig:gait_test_rough_rmse}.

The main conclusion which can be drawn from this study is that the best performance is obtained using the clustering trained on the mixed gait, as expected.  However, the slow gait clustering generalizes much better than the fast gait clustering.  The fixed-threshold base state estimator (denoted BSE) also performs quite well for this task, however because this was a walk-in-place there was mainly foot rotation and minimal slip; as seen from other tests, the clustering-based base state estimator performs much better when slip occurs.  Further investigation into the effect of training data gait is left to future work.

\subsection{IMUs for Clustering Versus Estimation}
\label{sec:imu_vs_no_imu_rmse}

\begin{figure}[h!]
\includegraphics[width=0.49\textwidth]{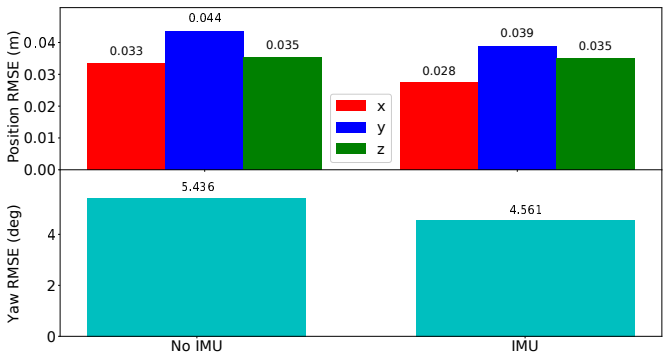}
\captionsetup{width=0.5\textwidth}
\caption{Root Mean Squared Error (RMSE) for estimation of the unobservable base position (top) and yaw (bottom) with and without using IMU data for online contact estimation.}
\label{fig:imu_vs_no_imu_rmse}
\end{figure}

As motivated in Sec. \ref{sec:sensing_for_clustering}, the use of endeffector IMU data in addition to contact wrench data essentially supervises the clustering problem, since the accelerometer and gyroscope capture linear and rotational slip.  We expect this sensor data to embed structure in the resulting clusters, meaning that IMUs should not be required when running the contact estimator afterwards.  To test this, we cluster using data points as in Eq. (\ref{eq:cluster_data}) but perform clustering-based state estimation with both a) the full data points including IMUs and b) without IMUs (dropping the last portion of Eq. (\ref{eq:cluster_data})). The resulting estimation errors are shown in Fig. \ref{fig:imu_vs_no_imu_rmse}.

Although the RMSE is slightly lower in all dimensions when using the IMU data, performance is not considerably changed when it is removed.  This is a very useful property because it means that the endeffector IMUs can be removed after initially collecting data for clustering.  While some robots are designed with endeffector IMUs, most are not; using this clustering method would involve temporarily attaching IMUs as in \cite{rotella_inertial_2016}, \cite{xinjilefu_imus_2016}.  This is reasonable for training, however attaching these sensors permamently involves designing rigid mounts, routing cables and protecting them from collisions with the environment.  The ability to remove IMUs after training the estimator is highly advantageous when working with real hardware.

\subsection{Clustering-Based Contact Estimation for Control}
\label{sec:clustering_for_control}

While the primary focus of this work has been on the development and evaluation of a contact probability estimator for use in base state estimation, the proposed method has applications in humanoid control as well.  The most direct application is the use of an improved base state estimator in a walking controller such as the one used to generate data in this work.  

This walking controller uses a simplified model in a model predictive control framework to plan center of mass and endeffector trajectories, which are tracked using an optimization-based inverse dynamics controller similar to \cite{herzog_momentum_2014}.  The estimated base pose is crucial in computing both dynamic model parameters and feedback control for endeffector tracking.  In the attached video, we demonstrate that the use of our contact probability estimator in this context improves control considerably, allowing the robot to walk on the rough terrain for a longer time before falling due to an accumulation of base state estimation error.

\section{CONCLUSIONS}

The clustering-based contact probability estimator presented in this work estimates the quality of contact using only proprioceptive sensor data in a completely unsupervised approach.  Unlike previous works, this estimator provides the probability of satisfying endeffector contact constraints in all six dimensions independently.  Use of this method in a base state estimation framework was shown to considerably lower estimation error as compared to a base state estimator which uses a fixed normal force threshold and noise parameters.  The proposed method also exhibits favorable properties which allow it to be used without endeffector IMUs after training and generalize to new terrain.  Finally, it was shown that use of this improved base state estimator for closed-loop inverse dynamics control allows the robot to remain stable during rough terrain walking for longer.  Future work will include further analysis of the properties of this contact estimator as well as a more low-level control application in which endeffector constraints in inverse dynamics are smoothly varied according to the contact probability.

\addtolength{\textheight}{-12cm}   % This command serves to balance the column lengths
                                  % on the last page of the document manually. It shortens
                                  % the textheight of the last page by a suitable amount.
                                  % This command does not take effect until the next page
                                  % so it should come on the page before the last. Make
                                  % sure that you do not shorten the textheight too much.

%====================================================================================================
%=========================================== Bibliography ===========================================

\bibliographystyle{IEEEtran} 
\bibliography{BibtexLibrary}

\end{document}